\def\year{2021}\relax
\documentclass[letterpaper]{article} 
\usepackage{aaai21}  
\usepackage{times}  
\usepackage{helvet} 
\usepackage{courier}  
\usepackage[hyphens]{url}  
\usepackage{graphicx} 
\usepackage{latexsym}
\usepackage{amsfonts}
\usepackage{amsmath}

\usepackage{natbib}  
\usepackage{caption} 
\title{Graph Information Vanishing Phenomenon in \\ Implicit Graph Neural Networks}
\author{
    Haifeng Li,
    Jun Cao,
    Jiawei Zhu,
    Qing Zhu,
    Guohua Wu

}
\affiliations{
 School of Geosciences and Info-Physics, Central South University, Changsha 410083, China
}

\begin{document}

\maketitle

\begin{abstract}
Graph Neural Networks (GNNs) have achieved great success in the field of graph representation learning by passing, transforming, and aggregating representations of neighbor nodes. One of the key problems of GNNs is how to describe the importance of neighbor nodes in the aggregation process for learning node representations. A class of GNNs solves this problem by learning implicit weights to represent the importance of neighbor nodes, which we call implicit GNNs such as Graph Attention Network. The basic idea of implicit GNNs is to introduce graph information with special properties followed by Learnable Transformation Structures (LTS) which encode the importance of neighbor nodes via a data-driven way. In this paper, we argue that LTS makes the special properties of graph information disappear during the learning process, resulting in graph information unhelpful for learning node representations. we call this phenomenon Graph Information Vanishing (GIV). Also, we find that LTS maps different graph information into highly similar results. To validate the above two points, we design two sets of 70 random experiments on five Implicit GNNs methods and seven benchmark datasets by using a random permutation operator to randomly disrupt the order of graph information and replacing graph information with random values. We find that randomization does not affect the model performance in 93\% of the cases, with about 7 percentage causing an average 0.5\% accuracy loss. And the cosine similarity of output results, generated by LTS mapping different graph information, over 99\% with an 81\% proportion. The experimental results provide evidences to support the existence of GIV in Implicit GNNs, and imply that the existing methods of Implicit GNNs do not make good use of graph information. The relationship between graph information and LTS should be rethought to ensure that graph information is used in node representation.

\end{abstract}

\section{Introduction}

GNNs have generalized deep learning from the Euclidean domain to the non-Euclidean domain and learn low-dimensional representations of nodes or graphs in an end-to-end fashion, making great success in many areas such as traffic network \cite{geng2019spatiotemporal}, recommendation systems \cite{wu2019session}, and computer vision \cite{gao2019know}. From the perspective of the spectral graph theory, \citeauthor{wu2020comprehensive} broadly classifies GNNs into two categories: spectral GNNs and spatial GNNs. Spectral GNNs \cite{bruna2013spectral,henaff2015deep,defferrard2016convolutional,kipf2016semi} need to perform eigenvalue decomposition on graph Laplacian matrix for convolution operations, thus are unsuitable for computing large-scale graphs and cannot adapt to graphs with different structures. Spatial GNNs \cite{duvenaud2015convolutional,velivckovic2017graph,hamilton2017inductive,ye2019curvature,zhao2020persistence} make GNNs more flexible and effective by defining convolution directly on the local structure of graph and aggregating neighbor node representations via a permutation invariant function. Most spatial GNNs can be integrated into Message Passing Neural Networks framework (MPNN) \cite{gilmer2017neural}, which contains a message-passing part and a readout part. The message-passing part regards node representations as messages and extracts localized information by transforming, aggregating and passing node representations \cite{fey2019fast}.

The performance of spatial GNNs depends heavily on the ability to measure the importance of messages in the aggregation process. The importance of messages is often interpreted as the weight of messages. We divide spatial GNNs into explicit GNNs and implicit GNNs based on whether the weight of messages needs to be learned. Explicit GNNs directly calculate the weight of messages from the graph information, with no need for learning. For example, GraphSAGE \cite{hamilton2017inductive} set the weight to 1 and GCN \cite{kipf2016semi} obtains it by decomposing the Laplace matrix of graph. Explicit GNNs often utilize simple graph information to identify the weight which cannot automatically adapt to datasets, while implicit GNNs introduce various graph information and design different learnable transformation structures (LTS), in order to overcome the defects of Explicit GNNs. For instance, GAT \cite{velivckovic2017graph} uses the hidden representation of nodes and a self-attention mechanism to learn the weight of messages. The self-attention mechanism is considered as the counterpart of the LTS in GAT. The LTS refers specifically to the structure which transforms graph information to obtain the weight of messages.

Current works usually consider that the role of LTS is to simply help graph information adapt to datasets and to preserve the particular properties of graph information. To test the idea, we elaborately select five implicit GNNs: CurvGN \cite{ye2019curvature}, PEGN \cite{zhao2020persistence}, GAT \cite{velivckovic2017graph}, HGCN \cite{chami2019hyperbolic}, and AGNN \cite{thekumparampil2018attention}. Then we analyze the performance of the models on seven node classification benchmark datasets, when we use a random permutation operator to randomly disrupt the order of graph information and replace the origin graph information with randomly sampled values from a uniform distribution. We surprisingly find that the numerical experiments do not support the above view: the two randomization operations do not cause significant degradation of accuracy of these models. Then we observe that LTS can transform different graph information into highly similar the weight of messages under the measure of visualization and cosine similarity. The experimental results indicate that LTS makes the special properties of graph information disappear during training, resulting in graph information unhelpful for learning node representations, which we called Graph Information Vanishing (GIV). Finally, we illustrate GIV only exists in implicit GNNs while not in Explicit GNNs. This paper makes the following contributions:
\begin{itemize}
\item We define the concept of graph information vanishing, which illustrates the specific property of graph information does not be preserved by Implicit GNNs. The relationship between graph information and LTS should be rethought to ensure that graph information is used in GNNs.
\item We show that LTS is forced to transform different inputs into highly similar outputs which all make the performance of models excellent.
\item The phenomenon of GIV does not exist in Explicit GNNs.
\end{itemize}

\section{Related Work}
GNNs, as black box models, arouse wide concern about its power and limits. \citeauthor{li2018deeper} explains the mechanism and over-smoothing problem of GCN by considering the convolution layer as symmetric Laplacian smoothing. \citeauthor{xu2018powerful} proposes a theoretical framework for analyzing the discriminative power of GNNs to distinguish different graph structures. \citeauthor{oono2019graph} also accounts for losing the expressive power of GCN and proposes a weight normalization for alleviating the problem. \citeauthor{barcelo2019logical} further analyzes the expressive power of GNNs for boolean node classification and adds readout functions to increase the logical expressiveness. Besides, some methods contribute to modifying network architectures to improve performance of GNNs. \citeauthor{wu2019session} simplifies graph convolutional networks to adapt large scale graph by removing nonlinearities weight matrices. \citeauthor{li2019deepgcns} expands the layers of GCN from 2 to 56 layers by referring to the concept of residual/dense connections in CNN. For graph classification task, \citeauthor{knyazev2019understanding} analyzes the effect of attention on the readout phase and proposes a weakly-supervised method to train attention. Note that the most of analysis works focus on GCN and its variants, which are regarded as Explicit GNNs in this paper. By contrast, the analysis of Implicit GNNs is not nearly enough.

\section{Method}
In this part, we firstly review some key components of GNNs, then reorganize the pipeline of implicit GNNs for future analysis. 

\subsection{Some key components of GNNs} GNNs consist of message passing and readout in Message Passing Neural Networks framework. In this paper, we only focus on message passing. The forward propagation formula of message passing can be summarized as follows:

\begin{equation}
h_{i}^{(l)}=\Upsilon^{(l)}\left(h_{i}^{(l-1)}, \Box_{j \in \mathcal{N}(i)} \Phi^{(l)}\left(h_{i}^{(l-1)}, h_{j}^{(l-1)}, e_{i, j}^{(l-1)}\right)\right)
\end{equation}
Where $h_{i}^{(l-1)} \in \mathbb{R}^{F}$ is the representation of the node $i$ on the $l-1$ layer, and $e_{i, j} \in \mathbb{R}^{D}$ is edge feature from node $j$ to node $i$, $\mathcal{N}(i)$ is the neighboring nodes of node $i$, $\Upsilon$ and $\Phi$ are differentiable functions, $\Box$ is a differentiable, permutation invariant aggregation function, e.g., sum, mean or max. In this paper, edge feature $e_{i, j}$ is interpreted as graph information which helps GNNs assign the weight of messages in aggregation. The forward propagation formula of most spatial GNNs, which is simplified as the aggregation and reweight parts, can be summarized as follows:

\begin{equation}
h_{i}^{(l)}= \sigma^{(l)}\left(\square_{j \in \overline{\mathcal{N}}(i)}\left(\tau_{i, j}^{(l)} W^{(l)} h_{j}^{(l-1)}\right)\right)
\end{equation}

\begin{equation}
\tau_{i, j}^{(l)}=\Delta^{(l-1)}\left(e_{i, j}^{(l-1)}\right)
\end{equation}
Where $\sigma$ is the activate function, $W$ is a matrix of filter parameters, $\tau_{i, j}$ is the weight of the message from node $j$ to node $i$, $\Delta$ is a transformation function.

If we need to learn the transformation function, such as multilayer perceptron (MLP), the spatial GNN is the implicit GNN, and vice versa is Explicit GNN. In other words, GNNs can be classified as Explicit GNNs or Implicit GNNs based on whether learning is needed to get the weight of messages.

\begin{figure}[t]
\centering
\includegraphics[width=0.9\columnwidth]{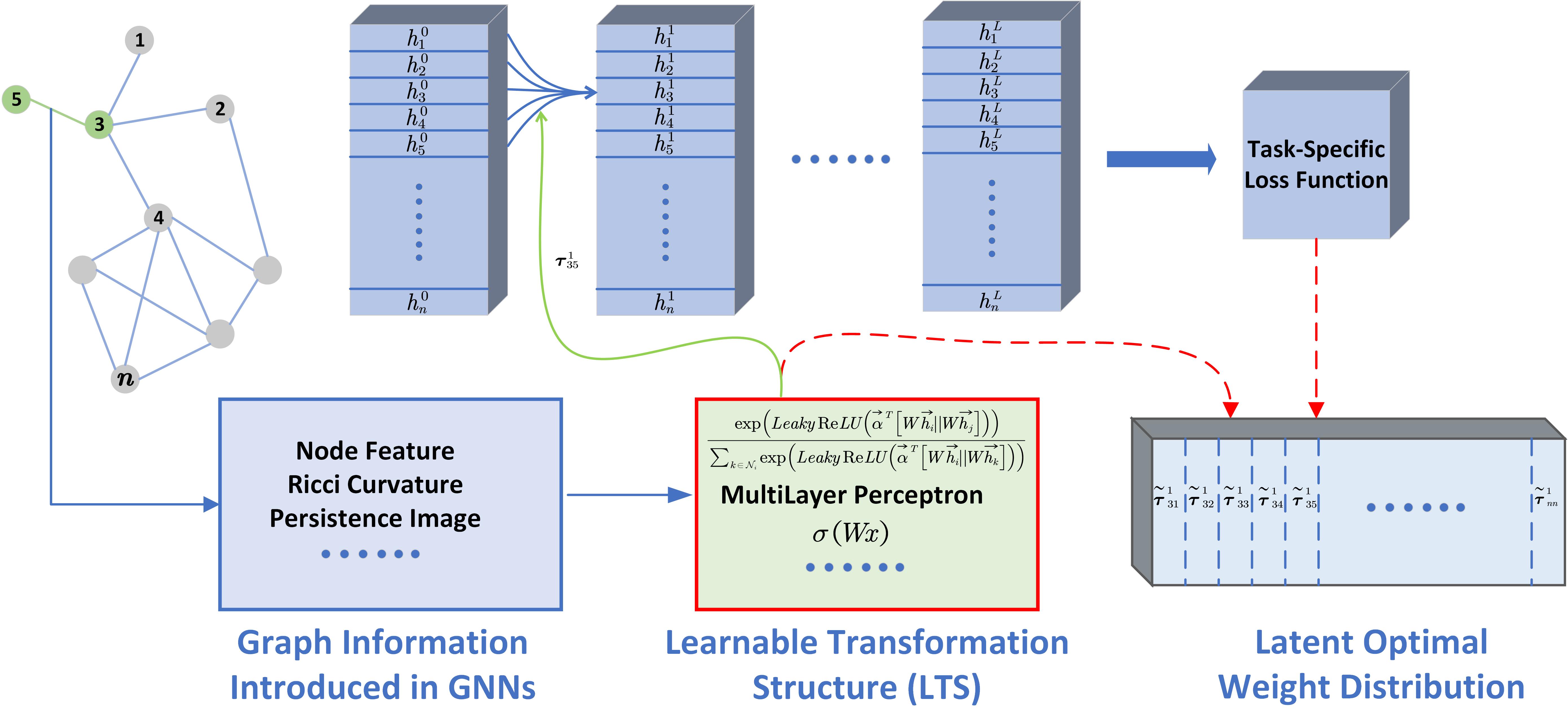} 
\caption{A pipeline of implicit GNNs. The upper part of the figure is the aggregation of GNNs, while the bottom is the reweight part. Given dataset and network architecture, we assume the task-specific loss function forces the LTS to learn the latently optimal weight distribution during training.}
\label{fig2}
\end{figure}

\subsection{The pipeline of implicit GNNs} For implicit GNNs, we define the learnable part of $\Delta$ as LTS particularly. A common assumption behind Implicit GNNs is that graph information $e_{i, j}$ can help GNNs improve performance by learning more knowledge about graph and function $\Delta$ helps graph information $e_{i, j}$ adapt to GNN architecture and dataset. Figure \ref{fig2} shows the general pipeline of Implicit GNNs, and the role of LTS is illustrated in the bottom of Figure \ref{fig2}. We  further show that some famous GNNs can be resolved into the pipeline.

\subsection{Implicit GNNs models}
Based on the pipeline of implicit GNNs defined above, we can rederive five Implicit GNNs: CurvGN, PEGN, GAT, HGCN, and AGNNN. The benefit of this statute is to helps us unify our analysis and understanding of the common problems which implicit GNNs faced with.

The five Implicit GNNs can be roughly divided into two categories according to the type of graph information: one is CurvGN and PEGN using additional ricci curvature and persistence images as graph information; the other is GAT, HGCN and AGNN constructing graph information through the hidden representation of nodes. They are classified as shown in Table \ref{tbl:1}.

\begin{table}[h]
\centering
\begin{tabular}{lll} 
\hline
Model  & Graph
  Information ($e_{i, j}$) & LTS($\Delta$)           \\ 
\hline
CurvGN & Ricci
  Curvature    & MLP                                      \\
PEGN   & Persistence
  Image       & MLP                                      \\ 
\hline
GAT    & $\left(h_{i} \| h_{j}\right)$ & $\vec{a}^{T} W$ \\
HGCN   & $\left(\log ^{K}\left(h_{i}\right) \| \log ^{K}\left(h_{j}\right)\right)$ & MLP                                      \\
AGNN   & $\cos \left(h_{i} \cdot h_{j}\right)$                                     & $\beta$                 \\
\hline
\end{tabular}
\caption{Summary of graph information and LTS for the five Implicit GNNs.}
\label{tbl:1}
\end{table}

\subsubsection{CurvGN} CurvGN thinks that ricci curvature endows GNNs with more discriminative power, because ricci curvature is a good measure of the connection strength between two adjacent nodes. Neighbors of adjacent nodes in the same community often have many shortcuts and large overlap. The curvature of corresponding edges is positive and information between the nodes is easy to interact. Nevertheless, the curvature of edges connecting two communities is negative and passing information is hard.

CurvGN considers ricci curvature as graph information $e_{i,j}$, and utilize two-layer MLP as the LTS. Then the formula of reweight part is:

\begin{equation}
    \tau_{i, j}^{(l)}=\operatorname{SOFTMAX}_{j \in \mathcal{N}(i)}\left(\operatorname{MLP}^{(l)}\left(e_{i, j}\right)\right)
\end{equation}

where $\operatorname{SOFTMAX}_{j \in \mathcal{N}(i)}$ indicates normalizing the weight to avoid a numerical explosion. CurvGN propose the scalar form and the vector form of the weight $\tau$. For the vector form, the dimension of $\tau ^{(l)}$ is the same as $h^{(l)}$. So CurvGN has the power to reweight each channel of messages. The formula of aggregation part is:
\begin{equation}
h_{i}^{(l)}=\sigma\left(\sum_{j \in \mathcal{N}(i)} \operatorname{diag}\left(\tau_{i, j}^{(l)}\right) W^{(l)} h_{j}^{(l-1)}\right)
\end{equation}
 
\subsubsection{PEGN} PEGN views local structural information of graph can improve the adaptability of GNN adapting to large graphs with heterogeneous topology. PEGN uses persistence homology, a principled mathematical tool,to characterize the loopiness of nodes' neighbors, which measures the information transmission efficiency of each node. And PEGN utilizes persistence images to quantitatively describe persistence homology of each edges. 

PEGN refers to persistence images of graph as graph information and selects two-layer MPL as its own LTS. Besides, the reweight part and the aggregation part of PEGN is same as CurvGN.

\subsubsection{GAT} GAT is probably the best known implicit GNN in GNNs domain. GAT transforms the hidden representation of nodes to attention coefficient by self-attention mechanism. And the attention coefficient is used for implicit assigning the weight of messages.

The graph information of GAT can be viewed as the concatenation of the hidden representation of adjacent nodes,$e_{i, j}=\left(h_{i}|| h_{j}\right), e_{i, j} \in \mathbb{R}^{2 F^{\prime}}$, and $||$ indicates the concatenation. Then GAT uses self-attention mechanism to transform graph information into the weight of messages. The formula of reweight part is:
\begin{equation}
\tau_{i, j}=\operatorname{SOFTMAX}_{j \in \mathcal{N}(i)}\left(\vec{a}^{T} W e_{i, j}\right)
\end{equation}
 
where the weight matrix $W \in \mathbb{R}^{2 F^{\prime} \times 2 F^{\prime}}$ and the weight vector $\vec{a} \in \mathbb{R}^{2 F^{\prime}}$ are shared by all information. So, the LTS of GAT is $\vec{a}^{T} W$. To ensure stability of training, GAT also utilize the K-head attention mechanism. The formula of aggregation part is:
\begin{equation}
h_{i}^{(l)}=\sigma\left(\sum_{j \in N(i)} \prod_{k=1}^{K}\left(\tau_{i, j}^{(k)} W^{(k)} h_{j}^{(l-1)}\right)\right)
\end{equation}

\subsubsection{HGCN} HGCN is one of the first methods to extend the representation of nodes from Euclidean space to hyperbolic space, and tries to minimize the distortion that occurs when embedding graphs with hierarchical structures. By taking an exponential map, HGCN maps the node features to the hyperboloid manifold. Logarithmic map is used to project the embedding vectors to the tangent space. As the tangent space can be considered as a Euclidean space and it is isomorphic to $\mathbb{R}^{d}$, all calculations in the aggregation process are in the tangent space. The output of aggregation is projected back to hyperbolic space via exponential map, in order to make the representation of nodes possess special properties of hyperbolic space.

HGCN uses the connection of the hidden representation of two nodes in the tangent space, i.e., $e_{i, j}=\left(\log ^{K}\left(h_{i}\right) \| \log ^{K}\left(h_{j}\right)\right)$, as graph information, where $K$ denotes the curvature of hyperbolic space. To transform graph information to weight of messages, MLP is used as the LTS. So, the formula of reweight part is:
\begin{equation}
\tau_{i, j}^{(l)}=\operatorname{SOFTMAX}_{j \in \mathcal{N}(i)}\left(\operatorname{MLP}^{(l)}\left(\mathbf{e}_{i, j}\right)\right)
\end{equation}

where $\tau$ is a scalar. The aggregation process in HGCN is very similar to that of GCN under MPNN framework and can be expressed as:
\begin{equation}h_{i}^{(l)}=\exp ^{K}\left(\sum_{j \in \overline{\mathcal{N}}(i)} \tau_{i, j}^{(l)} \log ^{K}\left(h_{j}^{(l-1)}\right)\right)\end{equation}
Note that some operations, such as the activation function for hyperbolic spaces, have been omitted to highlight the key parts of HGCN.

\subsubsection{AGNN} AGNN is a special kind of spatial GNN, which abandons the weight matrix of aggregation and only uses an attention propagation matrix to aggregate the hidden representation of nodes. The attention propagation matrix is generated by a special attention mechanism in a data-driven mode. And AGNN can learn dynamic and adaptive weight of messages in order to gain more accurate predictions.

AGNN refers to the cosine of the hidden representation of two adjacent nodes as graph information, 
$e_{i, j}=\cos \left(h_{i} \cdot h_{j}\right) \in \mathbb{R}$, and $\cos \left(h_{i} \cdot h_{j}\right)=h_{i}^{T} h_{j} /\left\|h_{i}\right\|\left\|h_{j}\right\|$. Then, the attention mechanism of AGNN is the reweight part, of which the formula is:
\begin{equation}
\tau_{i, j}^{(l)}=\operatorname{SOFTMAX}_{j \in \mathcal{N}(i)}\left(\beta^{(l)} e_{i, j}\right)
\end{equation}
where $\beta^{(l)}$ is a learnable parameter. So, the LTS of AGNN is the parameter $\beta$. If node $i$ and node $j$ are not adjacent, the corresponding item of attention propagation matrix is 0. According to MPNN framework,The formula of aggregation part is:
\begin{equation}
h_{i}^{(l)}=\sum_{j \in \mathcal{N}(i)} \tau_{i, j}^{(l)} h_{j}^{(l-1)}
\end{equation}
Besides, the first layer of AGNN is a fully connected layer to generate low-dimensional representation of nodes. The last layer of AGNN is also a fully connected layer aiming for outputting task-specific representation of nodes. Note that the two layers do not include the aggregation part, thus do not belong to graph convolution layer.

\subsection{Randomization Operations}
To verify the existence of GIV in Implicit GNNs, we design two randomization operations to change graph information. The details of the two operators will be introduced below.

\subsubsection{Random Permutation Operator} The purpose of the random permutation operator is to randomly disrupt the order of the graph information used by implicit GNNs. Thus, there is no longer a one-to-one correspondence between $e_{i,j}$, $\tau_{i,j}$ as shown in the upper part of Figure \ref{fig3}. The random permutation operator only disrupts the order in which the graph information corresponds to edges and does not change the values of graph information. We refer to the GNNs containing a random permutation operator as random permutation GNNs.

\subsubsection{Random Values Substitution} 
\begin{figure}[htbp]
\centering
\includegraphics[width=0.9\columnwidth]{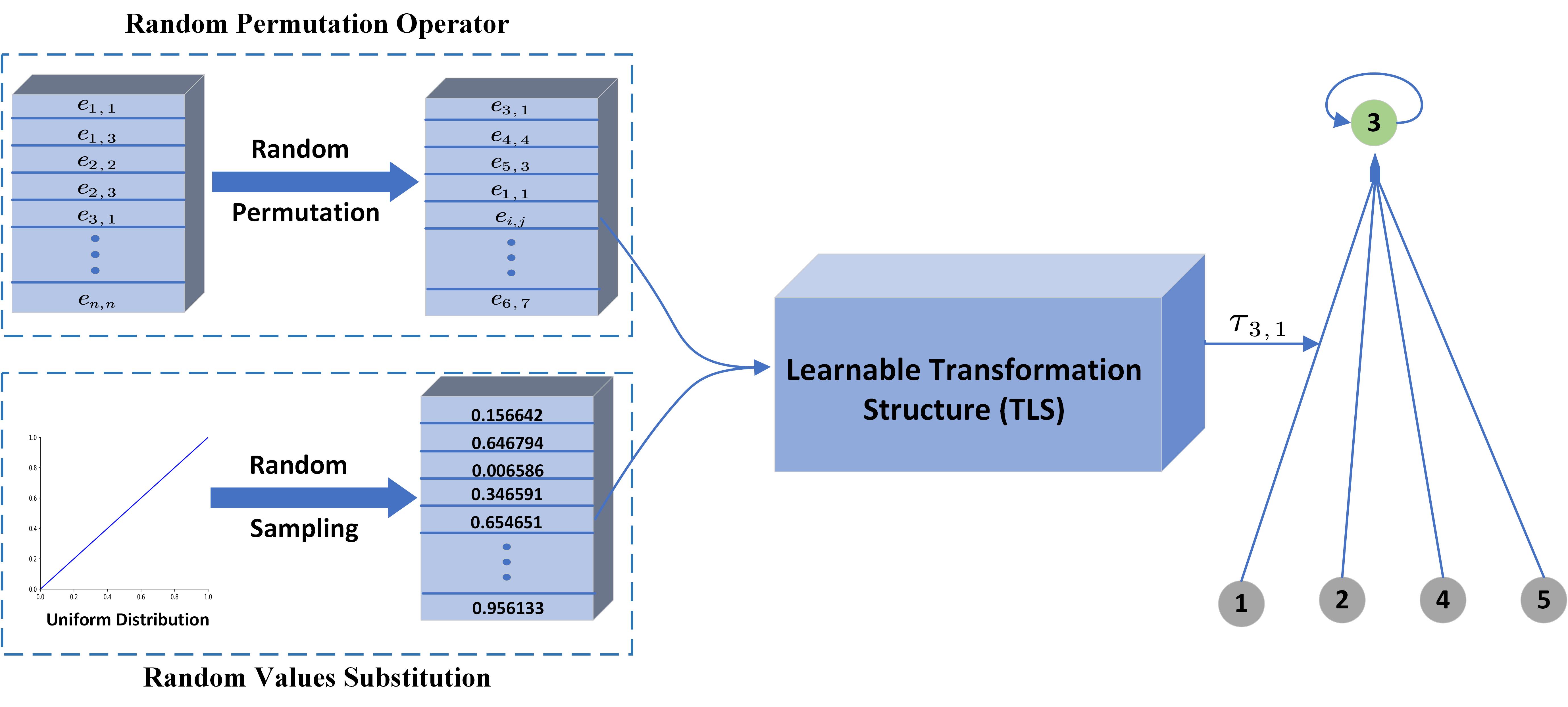} 
\caption{An illustration of the two randomization operators. The weight   is computed by input different random values into LTS.}
\label{fig3}
\end{figure}

Random value substitution is proposed to replace the graph information used by implicit GNNs with random values. The operator can strongly verify whether the specific properties of graph information are helpful for GNNs, because random values lack of the special properties of graph information. Random value substitution generates random values by randomly sampling a probability distribution, as shown in the lower part of Figure \ref{fig3}. GNNs with random value substitution are considered as RandGNNs.

\section{Experiments and Results}
\begin{table*}[h]
\centering
\begin{tabular}{llllllll} 
\hline
Methods     & Cora   & Citeseer & PubMed & \begin{tabular}[c]{@{}l@{}} Coauthor \\CS \end{tabular} & \begin{tabular}[c]{@{}l@{}} Coauthor \\Physics \end{tabular} & \begin{tabular}[c]{@{}l@{}} Amazon \\Computers \end{tabular} & \begin{tabular}[c]{@{}l@{}} Amazon\\~Photo \end{tabular}  \\ 
\hline
CurvGN               & 82.16$\pm$0.52            & 71.78$\pm$0.72            & \textbf{79.02$\pm$0.43}   & \textbf{92.42$\pm$0.34} & 93.39$\pm$0.25            & 83.91$\pm$0.45            & 91.15$\pm$0.61             \\
 CurvGN\_A\textbf{} & {82.12$\pm$0.67} & {71.74$\pm$0.71} & 78.96$\pm$0.33             & 92.23$\pm$0.39            & \textbf{93.42$\pm$0.26} & 84.11$\pm$0.47            & \textbf{91.35$\pm$0.59}  \\
CurvGN\_B            &\textbf{82.43}$\pm$0.49            & \textbf{71.85}$\pm$0.60            & 78.64$\pm$0.40             & 92.33$\pm$0.43            & 93.33$\pm$0.29            & \textbf{84.20$\pm$0.55} & 91.14$\pm$0.54             \\
CurvGN\_C            & 82.08$\pm$0.52            & 71.71$\pm$0.62            & 78.94$\pm$0.43             & 92.31$\pm$0.29            & 93.38$\pm$0.32            & 83.99$\pm$0.58            & 91.32$\pm$0.55             \\ 
\hline
PEGN                 & 82.12$\pm$0.61            & 71.72$\pm$0.64            & \textbf{78.98$\pm$0.32}   & 92.46$\pm$0.42            & 93.29$\pm$0.29            & 82.31$\pm$0.80            & \textbf{91.16$\pm$0.59}  \\
PEGN\_A              & 82.10$\pm$0.63            & 71.64$\pm$0.61            & 78.91$\pm$0.34             & 92.30$\pm$0.40            & \textbf{93.44$\pm$0.30} & \textbf{83.57$\pm$0.74} & 91.02$\pm$0.59             \\
PEGN\_B              & 82.29$\pm$0.77            & 71.62$\pm$0.67            & 78.94$\pm$0.30             & \textbf{92.46$\pm$0.34} & 93.35$\pm$0.29            & 83.24$\pm$0.73            & 91.09$\pm$0.50             \\
PEGN\_C              & \textbf{82.32$\pm$0.53}  & \textbf{71.73$\pm$0.57}  & 78.94$\pm$0.39             & 92.34$\pm$0.39            & 93.31$\pm$0.30            & 83.36$\pm$0.62            & 90.98$\pm$0.54             \\ 
\hline
GAT                  & 82.55$\pm$0.75            & 71.58$\pm$0.70            & 77.55$\pm$0.50             & \textbf{91.20$\pm$0.51} & \textbf{92.33$\pm$0.69} & \textbf{82.75$\pm$0.98} & \textbf{91.83$\pm$0.87}  \\
GAT\_A               & 82.61$\pm$0.78            & \textbf{71.62$\pm$0.81}  & 77.63$\pm$0.50             & 90.95$\pm$0.36            & 92.20$\pm$0.64            & 82.36$\pm$0.91            & 91.25$\pm$0.79             \\
GAT\_B               & \textbf{82.62$\pm$0.81}  & 71.47$\pm$0.74            & \textbf{77.63$\pm$0.48}  & 90.98$\pm$0.46            & 92.21$\pm$0.60            & 82.40$\pm$0.93            & 91.21$\pm$0.94             \\
GAT\_C               & 82.54$\pm$0.75            & 71.55$\pm$0.76            & 77.48$\pm$0.51             & 90.95$\pm$0.38            & 92.24$\pm$0.62            & 82.69$\pm$0.83            & 91.13$\pm$0.80             \\ 
\hline
HGCN                 & 80.18$\pm$1.47            & 67.89$\pm$1.31            & 76.99$\pm$0.79             & 91.31$\pm$0.55            & N\textbackslash{}A    & \textbf{81.71$\pm$1.29} & \textbf{91.22$\pm$0.83}  \\
HGCN\_A              & \textbf{80.39$\pm$1.18}  & 67.88$\pm$1.60            & 77.25$\pm$0.63             & 91.40$\pm$0.62            & N\textbackslash{}A    & 81.36$\pm$1.25            & 90.58$\pm$0.82             \\
HGCN\_B              & 80.16$\pm$1.28            & \textbf{68.14$\pm$1.34} & \textbf{77.26$\pm$0.76}  & \textbf{91.54$\pm$0.62} & N\textbackslash{}A    & 81.31$\pm$1.28            & 90.41$\pm$0.90             \\
HGCN\_C              & 80.19$\pm$1.32            & 68.09$\pm$1.38            & 77.07$\pm$0.79             & 91.50$\pm$0.49            & N\textbackslash{}A    & 81.25$\pm$1.21            & 90.63$\pm$0.79             \\ 
\hline
AGNN                 & 81.29$\pm$1.01            & 70.67$\pm$0.99            & 77.44$\pm$0.61             & \textbf{90.20$\pm$0.52} & 93.00$\pm$0.28            & \textbf{77.48$\pm$2.92} & \textbf{89.96$\pm$0.98}  \\
AGNN\_A              & \textbf{81.47$\pm$0.84}  & 70.73$\pm$1.27            & \textbf{78.09$\pm$0.41}  & 90.11$\pm$0.49            & 93.27$\pm$0.35            & 77.03$\pm$1.52            & 89.75$\pm$0.77             \\
AGNN\_B              & 81.42$\pm$0.96            & 70.55$\pm$1.22            & 76.93$\pm$0.60             & 89.80$\pm$0.59            & 93.04$\pm$0.30            & 77.20$\pm$1.56            & 89.50$\pm$0.78             \\
AGNN\_C              & 81.29$\pm$1.19            & \textbf{70.92$\pm$1.08}  & 77.18$\pm$0.54             & 90.00$\pm$0.53            & \textbf{93.30$\pm$0.33} & 77.38$\pm$1.54            & 89.87$\pm$0.61             \\
\hline
\end{tabular}
\caption{Summary of statistic results in term of comparing different implicit GNNs with the corresponding random permutation models for seven benchmark datasets.}
\label{tbl:3}
\end{table*}
In this paper, random permutation operator and random value substitution are utilized to understand the existence of GIV in Implicit GNNs. To verify that graph information is not correlated with the weight of messages, we use a random permutation operator to disrupt the order of graph information and compare random permutation GNNs with the original models. Further, we utilize random value substitution to show that the LTS does not maintain the properties of graph information during training. In order to verify that the LTS transforms different inputs into highly similar outputs, we qualitatively and quantitatively depict the similarity of the weight of messages generated by the LTS with different graph information.

\subsection{Experimental Settings}
We select seven benchmark datasets: Cora, Citeseer, PubMed, Coauthor CS, Coauthor Physics, Amazon Computers, and Amaz on Photos. The training set consists of 20 nodes per class, the validate set and the test set include 500 and 1000 nodes respectively. We consider Cora, Citeseer, and PubMed as sparse datasets due to relatively small average node degree, while the other four datasets are dense.

We introduce how to perform random operations. 0, 10 and 100 are chosen as the seeds of random permutation operator, and their corresponding models are named as model\_A, model\_B, and model\_C, respectively. We use 0-1 uniform distribution as the probability distribution of random permutation operator, and choose 2020 and 2021 as the random seeds. The 5 implicit GNNs with random permutation operator are named as RandCurvGN, RandPEGN, RandGAT, RandHGCN and RandAGNN, respectively. To avoid the effect of random initialization of weight matrices, the averaged weights of messages obtained from 50 times repeated experiments are the final weight. We choose GCN as a representative of explicit GNNs. We change implicit GNNs to explicit GNNs by taking the final weight as the weight of the models and removing the LTS. For GAT and CurvGN, their corresponding explicit GNNs are called GAT\_E and CurvGN\_E.

\begin{table*}[h]
\centering
\begin{tabular}{llllllll} 
\hline
Methods     & Cora   & Citeseer & PubMed & \begin{tabular}[c]{@{}l@{}} Coauthor \\CS \end{tabular} & \begin{tabular}[c]{@{}l@{}} Coauthor \\Physics \end{tabular} & \begin{tabular}[c]{@{}l@{}} Amazon \\Computers \end{tabular} & \begin{tabular}[c]{@{}l@{}} Amazon\\~Photo \end{tabular}  \\ 
\hline
CurvGN     & 82.16$\pm$0.52 &\textbf{71.78$\pm$0.72}& 79.02$\pm$0.43 & \textbf{92.42$\pm$0.34}  &\textbf{93.39$\pm$0.25}        & 83.91$\pm$0.45       & 91.15$\pm$0.61    \\
RandCurvGN & \textbf{82.27$\pm$0.48} & 71.55$\pm$0.68 & \textbf{79.07$\pm$0.37} & 92.27$\pm$0.39  & 93.39$\pm$0.26         & \textbf{84.11$\pm$0.50}       & \textbf{91.34$\pm$0.66}    \\ 
\hline
PEGN       & 82.12$\pm$0.61 & \textbf{71.72$\pm$0.64} & \textbf{78.98$\pm$0.32} & \textbf{92.46$\pm$0.42}  & 93.29$\pm$0.29         & 82.31$\pm$0.80       & 91.16$\pm$0.59    \\
RandPEGN   & \textbf{82.23$\pm$0.56}& 71.66$\pm$1.09 & 78.98$\pm$0.42 & 92.30$\pm$0.41  & \textbf{93.44$\pm$0.28}        & \textbf{84.04$\pm$0.58}      & \textbf{91.29$\pm$0.68}  \\ 
\hline
GAT        & 82.55$\pm$0.75 & 71.58$\pm$0.70 & 77.55$\pm$0.50 & \textbf{91.20$\pm$0.51}  & \textbf{92.33$\pm$0.69}         & 82.75$\pm$0.98       & \textbf{91.83$\pm$0.87}    \\
RandGAT    &\textbf{82.62$\pm$0.69} &\textbf{71.80$\pm$0.66} &\textbf{77.68$\pm$0.41} & 91.04$\pm$0.43  & 92.25$\pm$0.64         &\textbf{82.79$\pm$0.85}       & 91.57$\pm$0.64    \\ 
\hline
HGCN       & \textbf{80.18$\pm$1.47} & 67.89$\pm$1.31 & 76.99$\pm$0.79 & 91.31$\pm$0.55  & N\textbackslash{}A & \textbf{81.71$\pm$1.29}       & 91.22$\pm$0.83    \\ 
RandHGCN   & 80.17$\pm$1.34 & \textbf{68.04$\pm$1.52} & \textbf{77.09$\pm$0.81} & \textbf{91.48$\pm$0.58}  & N\textbackslash{}A & 81.67$\pm$1.12       & \textbf{91.29$\pm$0.64}    \\ 
\hline
AGNN       & \textbf{81.29$\pm$1.01} & 70.67$\pm$0.99 & \textbf{77.44$\pm$0.61} & \textbf{90.20$\pm$0.52}  & 93.00$\pm$0.28         & \textbf{77.48$\pm$2.92}       & \textbf{89.96$\pm$0.98}    \\
RandAGNN   & 80.76$\pm$0.86 & \textbf{70.91$\pm$1.26} & 76.13$\pm$0.69 & 88.81$\pm$0.81  & \textbf{93.13$\pm$0.34}         & 76.80$\pm$1.79       & 89.89$\pm$0.51    \\
\hline
\end{tabular}
\caption{Summary of statistic results in term of comparing different implicit GNNs with the corresponding RandGNNs for seven benchmark datasets.}
\label{tbl:4}
\end{table*}

\begin{table}
\small
\centering
\begin{tabular}{llll} 
\hline
Methods    & Cora                  & \begin{tabular}[c]{@{}l@{}}Coauthor\\Physics \end{tabular} & \begin{tabular}[c]{@{}l@{}}Amazon\\Photo \end{tabular}  \\ 
\hline
GAT        & 5.58$\times 10^{-4}$  & 1.34$\times 10^{-3}$                                       & 1.40$\times 10^{-3}$                                    \\
GAT\_      & 1.65$\times 10^{-4}$  & 1.61$\times 10^{-3}$                                       & 1.54$\times 10^{-3}$                                    \\
RandGAT    & 1.23$\times 10^{-2}$  & 1.73$\times 10^{-3}$                                       & 1.22$\times 10^{-3}$                                    \\ 
\hline
CurvGN     & 2.60$\times 10^{-4}$  & 2.05$\times 10^{-5}$                                       & 9.41$\times 10^{-6}$                                    \\
CurvGN\_   & 8.26$\times 10^{-5}$  & 1.08$\times 10^{-5}$                                       & 6.40$\times 10^{-7}$                                    \\
RandCurvGN & 9.98$\times 10^{-5}$  & 2.54$\times 10^{-5}$                                       & 4.18$\times 10^{-6}$                                    \\
\hline
\end{tabular}
\caption{Summary of consistency of hidden layers' weight of messages for different models.}
\label{tbl:5}
\end{table}

\subsection{Implicit GNNs with Random Permutation Operator}
In this section, we explore the effect of random permutation operator on implicit GNNs. Table \ref{tbl:3} shows the classification accuracy of five groups of models on seven datasets. For each group of models, the best results are roughly randomly distributed among model, model\_A. model\_B and model\_C. On most datasets, the difference between the best and the worst accuracy is within 0.5\%, which indicates that there is no one-to-one correspondence between graph information and the weight of messages.

We notice that on two dense datasets, that is, Amazon Computers and Amazon Photo, the accuracies of GAT, HGCN and AGNN are slightly higher than their corresponding randomly permutated model. If we add the random permutation operator to these models, the input of the LTS will be changed with iteration, which may lead to a more difficult optimization, due to the graph information consisting of the hidden representation of nodes. Even for Amazon Photo, which has the largest difference of accuracy, the accuracies of GAT and HGCN are only about 0.6\% higher than the second-best model, which means only 6 more nodes are predicted correctly considering that there are only 1000 nodes in the test set. Compared to the total number of nodes (7487) in the dataset, 6 nodes appear to be insignificant. Therefore, we believe that random permutation operator do not cause substantial difference to the five models.

\subsection{Random Graph Information}
We discuss the influence of random value substitution to implicit GNNs in this section. As shown in Table \ref{tbl:4}, the best results lie randomly on implicit GNNs and RandGNNs for five group of models on different datasets. The accuracy of each set in Table \ref{tbl:4} has a smaller variation range (smaller than 0.2\% on most datasets) compared to Table \ref{tbl:3}, which indicates that replacing the original information models depend on with random values hardly influence the performance of models. The 0-1 uniformly distributed sampling provides an appropriate randomly initialized input to the LTS.

On dense datasets, GAT correctly predicts 1 to 4 more nodes than RandGAT. We believe that it is because the model uses the hidden representation of nodes as information, and the weight matrix used to calculate node representation will indirectly increase the capability of the LTS, making it easier to approximate the latently optimal weight distribution. Since the LTS of each layer of AGNN has only one learnable parameter, its transformation ability is not sufficient and the situation described above is more obvious

\begin{table*}[h]
\centering
\begin{tabular}{lccccccc} 
\hline
Methods     & Cora   & Citeseer & PubMed & \begin{tabular}[c]{@{}l@{}} Coauthor \\CS \end{tabular} & \begin{tabular}[c]{@{}l@{}} Coauthor \\Physics \end{tabular} & \begin{tabular}[c]{@{}l@{}} Amazon \\Computers \end{tabular} & \begin{tabular}[c]{@{}l@{}} Amazon\\~Photo \end{tabular}  \\ 
\hline
GAT,
  RandGAT       & ~
  0.9997 & ~
  0.9999 & ~
  0.9998 & ~
  0.9975 & ~
  0.9982     & ~
  0.9380     & ~
  0.9279  \\ 

CurvGN,
  RandCurvGN & ~
  0.9998 & ~
  0.9989 & ~
  0.9993 & ~
  0.9999 & ~
  0.9999     & ~
  0.9999     & ~
  0.9999  \\ 

PEGN,
  RandPEGN     & ~
  1.0      & ~
  1.0      & ~
  1.0      & ~
  1.0      & ~
  1.0          & ~
  0.9865   & ~
  0.9715  \\
\hline
\end{tabular}
\caption{Summary of cosine similarity of hidden layer's weight of messages for LTS with different graph information. Keep only 4 decimal places and round off the rest.}
\label{tbl:6}
\end{table*}

\begin{figure}[htbp]
\centering
\includegraphics[width=0.9\columnwidth]{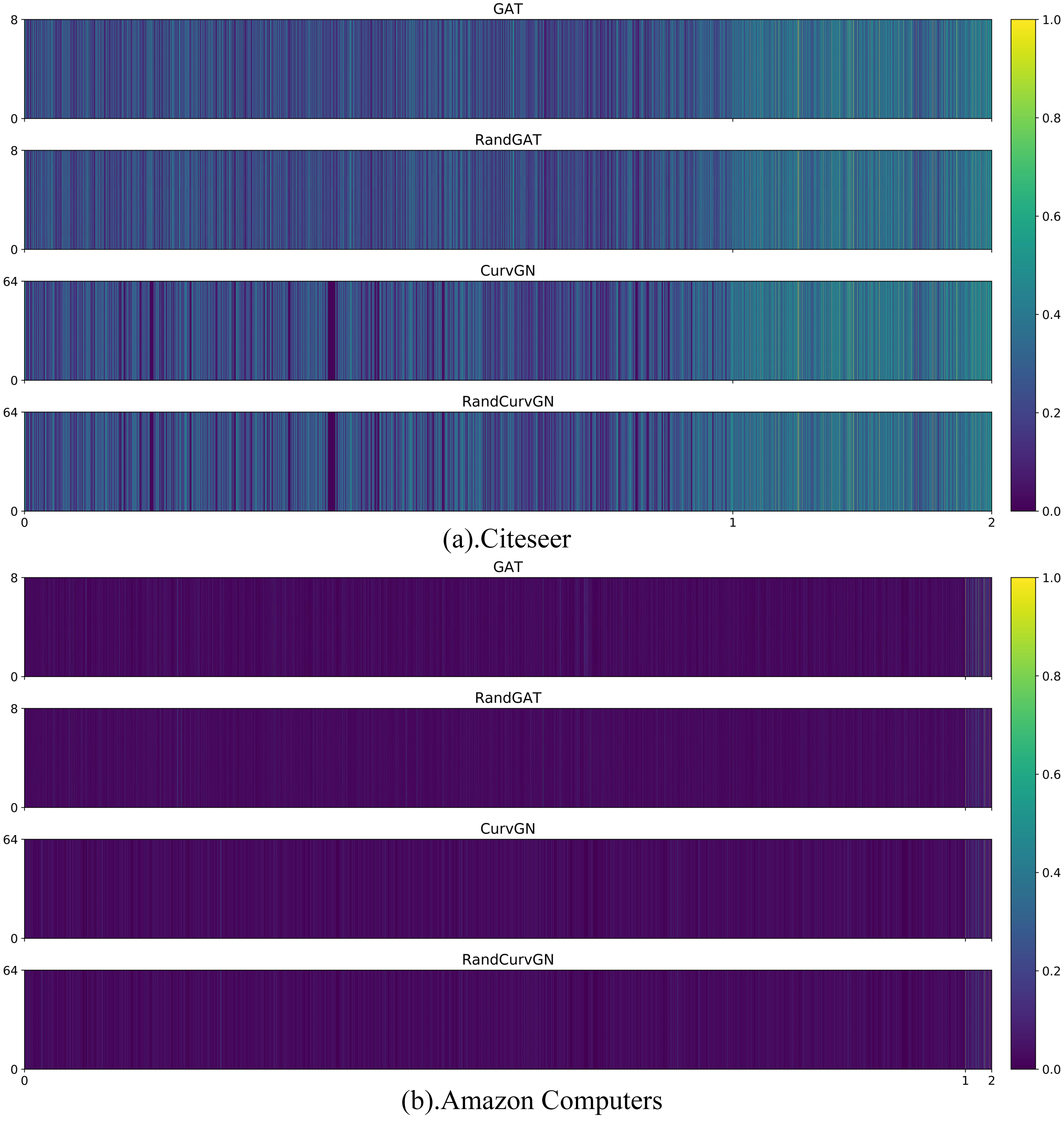} 
\caption{An illustration of visualization of hidden layer's weight of messages for different models on different datasets. (a) visualizes weight of messages on Citeseer, and (b) visualizes it on Amazon Computers. The vertical axis of each subfigure represents the dimension of weight of messages. The horizontal axis of the subfigure is divided into two parts: 0 to 1 for the weight of the edge and 1 to 2 for the weight of the self-loop.}
\label{fig4}
\end{figure}

\subsection{Similarity of Weight of Messages}
In this section, we focus on the similarity of weight of messages obtained by the LTS with different input from both qualitative and quantitative perspectives. Firstly, we analyzed the properties of weight of messages of two models, GAT and CurvGN. Figure \ref{fig4} visualizes the weight of messages in the hidden layer of GAT, RandGAT, CurvGN, and RandCurvGN on the sparse and dense datasets. Whether on sparse datasets or dense datasets, we can hardly observe any difference in the weight of messages between GAT and RandGAT. And the above phenomenon also exists in CurvGN. GAT and CurvGN have slight differences in the weight of messages which can be observed with the naked eye, due to their different network architectures. This qualitatively illustrates the ability of the LTS to transform different graph information into highly similar weight of messages, with little impact on the performance of models.

We use cosine similarity to quantitatively measure the similarity of weight of messages. Since the weight of messages may be a matrix, it needs to be converted to vector form. From Figure \ref{tbl:4}, we know that the value of each column of weight of messages changes slightly. To quantify the average magnitude of change in each column of weight of messages, we depict the following formula:

\begin{equation}C\left(A_{m \times n}\right)=\frac{\sum_{j}^{n} \sum_{i}^{m}\left|a_{i, j}-\frac{\sum_{k}^{m} a_{k, j}}{m}\right|}{m \times n}\end{equation}
where $A$ denotes weight of messages, $m$ is the dimension of the weight, and $n$ is the number of edges. As shown in Table \ref{tbl:5}, the average fluctuation range for each column of weight of messages is quite small and almost negligible. Therefore, the weight of messages in matrix form is transformed into vector form by averaging the columns, and the similarity of weight of messages is depicted by cosine similarity.

\begin{equation}\operatorname{cosineSIM}\left(\overrightarrow{A_{1}}, \overrightarrow{A_{2}}\right)=0.5 \frac{\overrightarrow{A_{1}} \cdot \overrightarrow{A_{2}}}{\left|\overrightarrow{A_{1}}\right| \times\left|\overrightarrow{A_{2}}\right|}+0.5\end{equation}
where $\overrightarrow{A}$ denotes the weight of messages in vector form.

Table \ref{tbl:6} shows the effect of different graph information on the similarity of weight of messages. From Table \ref{tbl:6}, we can see that in the case where the average node degree of the dataset is relatively small, for all three models, the cosine similarity is almost equal to 1 even if the weight of messages is obtained using different graph information. We observed a small decrease in the similarity of weights of messages for GAT and PEGN on Amazon Computers and Amazon Photo, which have the largest average node degree. This may be because GAT and PEGN use 128 and 50 dimensional vectors as graph information, respectively, and the LTS may not have sufficient capability to fit large scale and dense datasets. For GAT, since the weight matrix used to compute node representation may indirectly increase the capability of LTS, causing the similarity of weights of messages further decrease. It is illustrated that changing the inputs to the LTS has little or no effect on the model and explains the numerical results for Table \ref{tbl:3} and Table \ref{tbl:4}. It is also shown that the LTS erases the special properties of graph information during training, leading to GIVs for Implicit GNNs.

\begin{table}
\centering
\begin{tabular}{llll} 
\hline
Methods  & Cora                     & Citeseer                 & PubMed                    \\ 
\hline
GCN      & 81.44$\pm$0.89           & 70.71$\pm$0.83           & 78.61$\pm$0.52            \\
GCN\_A    & 67.12$\pm$1.96           & 58.84$\pm$1.77           & 71.32$\pm$1.38            \\
GCN\_B    & 69.07$\pm$1.70           & 54.27$\pm$1.78           & 72.67$\pm$1.09            \\
GCN\_C    & 67.44$\pm$1.34           & 58.02$\pm$1.55           & 73.23$\pm$2.28            \\
GCN\_1    & 77.21$\pm$1.37           & 65.14$\pm$1.30           & 77.49$\pm$0.70            \\ 
\hline
GAT\_E    & \textbf{82.66$\pm$0.63}  & 71.39$\pm$1.00           & 77.62$\pm$0.52            \\
CurvGN\_E & 82.18$\pm$0.54           & \textbf{72.01$\pm$0.61}  & \textbf{79.00$\pm$0.38}   \\
\hline
\end{tabular}
\caption{Summary of statistic results in term of different kinds of Explicit GNNs for seven benchmark datasets.}
\label{tbl:8}
\end{table}

\subsection{Analyzing Explicit GNNs}
In this section, we discuss whether GIV exists in Explicit GNNs. Table \ref{tbl:8} shows the effect of different weight of messages on the performance of Explicit GNNs. GCN\_1 indicates the weight of messages set to 1. By comparing different types of GCNs in Table \ref{tbl:8}, we find that for GCN, model performance is significantly degraded after randomly disrupting the order of graph information (degree of nodes). The performance of GCN is also significantly degraded if no specific weight is assigned to messages. This shows that proper weight of messages plays a key role in GNNs and the phenomenon of GIV does not exist in GCN. Further, we change implicit GNNs into explicit GNNs. The performance of GAT\_E and CurvGN\_E in Table \ref{tbl:8} is comparable to that of the corresponding GAT and CurvGN, and is substantially improved relative to GCN. This shows that the reason for the superior performance of Implicit GNNs is the more appropriate weight of messages obtained by learning, not the graph information.

\section{Discussion}
For CurvGN and PEGN, we find that neither random permutation operator nor random value substitution affect model performance on these seven datasets. Besides, the LTS with different inputs can output highly similar outputs. We call this phenomenon “graph information vanishing”. Graph information vanishing is also reflected in GAT, HGCN and AGNN. For the latter type of models, we still recommend using the hidden representation of nodes as the input of the LTS. There are two reasons: firstly, it doesn't cause additional computational cost; secondly, the weight matrix transforming node representation might indirectly increase the capability of the LTS.

Given model structure and dataset, the LTS will inevitably transform different inputs to approximate the latently optimal weight distribution to minimize the loss, since the loss function of calculating node features and weight of messages is the same one and is the cross entropy with respect to node features. This results in the high similarity of weight of messages obtained by learning, as shown in Figure \ref{fig4}.

We propose two ideas to deal with the problem of implicit GNNs: one is to design a sophisticated network structure to better approximate the latently optimal weight distribution; the other is to change the loss function of the LTS. For the first idea, no model has the best performance on all datasets from Table \ref{tbl:3} and Table \ref{tbl:4}, which suggests that we could try to design a new LTS or network structure with better generalization capabilities. For the second idea, we can change the loss function of the LTS, making the weight of messages learn the structural information of graphs. Inspired by network embeddings \cite{perozzi2014deepwalk, grover2016node2vec,tang2015line}, the loss function that retains the local or global topology structures may be added to the loss function of the LTS.

\section{Conclusion and Future Works}
In this paper, we delve into the relationship between graph information and LTS in Implicit GNNs. We conduct a series of experiments to illustrate that the random permutation operator and the random value substitution do not significantly affect the performance of the Implicit GNNs. We then find that the weight of messages output by the LTS with different graph information is highly similar from qualitative and quantitative perspectives. These experimental results show that the GIV phenomenon does exist in Implicit GNNs and that the special properties of graph information are not utilized by GNNs, which inspires us to explore new directions to help GNNs learn the rich knowledge behind graphs.


\bibliography{ref}

\end{document}